\documentclass[conference]{IEEEtran}
\IEEEoverridecommandlockouts

\usepackage{cite}
\usepackage{amsmath,amssymb,amsfonts}
\usepackage{algorithmic}
\usepackage{algorithm}
\usepackage{graphicx}
\usepackage{textcomp}
\usepackage{xcolor}
\usepackage{booktabs}
\usepackage{multirow}
\usepackage{url}
\usepackage[utf8]{inputenc}
\usepackage[T1]{fontenc}
\usepackage{microtype}
\usepackage[font=footnotesize,labelfont=bf]{caption}
\usepackage{subcaption}
\usepackage[none]{hyphenat}
\usepackage{hyperref}
\usepackage[letterpaper, top=0.75in, bottom=0.75in, left=0.75in, right=0.75in]{geometry}

\begin{document}

\title{\vspace{0.25in}How Well Do Vision-Language Models Understand Sequential Driving Scenes? A Sensitivity Study}

\author{\IEEEauthorblockN{Roberto Brusnicki, Mattia Piccinini, Johannes Betz}
\IEEEauthorblockA{\textit{Professorship of Autonomous Vehicle Systems} \\
\textit{TUM School of Engineering and Design, Technical University of Munich}\\
Munich, Germany \\
\{roberto.brusnicki, mattia.piccinini, johannes.betz\}@tum.de}
}

\maketitle

\begin{abstract}
    Vision-Language Models (VLMs) are increasingly proposed for autonomous driving tasks, yet their performance on sequential driving scenes remains poorly characterized, particularly regarding how input configurations affect their capabilities. We introduce VENUSS (VLM Evaluation oN Understanding Sequential Scenes), a framework for systematic sensitivity analysis of VLM performance on sequential driving scenes, establishing baselines for future research. Building upon existing datasets, VENUSS extracts temporal sequences from driving videos, and generates structured evaluations across custom categories. By comparing 25+ existing VLMs across 2,600+ scenarios, we reveal how even top models achieve only 57\% accuracy, not matching human performance under similar constraints (65\%) and exposing significant capability gaps. 
    Our analysis shows that VLMs excel with static object detection
    but struggle with understanding vehicle dynamics and temporal relations.
    VENUSS offers the first systematic sensitivity analysis of VLMs focused on how input image configurations -- resolution, frame count, temporal intervals, spatial layouts, and presentation modes -- affect performance on sequential driving scenes.
    Supplementary material available at \url{https://TUM-AVS.github.io/VENUSS/}.
\end{abstract}

\section{Introduction}
\label{sec:introduction}

Vision-Language Models (VLMs) are increasingly proposed for autonomous driving (AD) tasks, from scene understanding to decision-making, by integrating visual and textual information \cite{Gao2026, pan2024vlp, li2025applications, yin2024survey, feng2025verdi, ma2024lampilot}. Pre-trained on extensive image-text datasets, VLMs have the potential to generalize in a zero-shot manner and handle rare events, providing interpretable explanations in natural language \cite{yin2024survey, li2025applications, feng2025verdi, jiang2025alphadrive, emvlm4ad2024, atakishiyev2024explainable}.

However, a critical gap persists between these promising claims and actual VLM performance on basic driving scene understanding. While existing benchmarks like VLADBench \cite{li2025finegrained} provide fine-grained evaluation, many remain "insufficient to assess capabilities in complex driving scenarios" under diverse conditions. A systematic \textit{temporal gap} exists where the VLMs' ability to interpret temporal changes (acceleration, deceleration, or directional shifts) is inadequately assessed \cite{ko2025st}. Moreover, existing datasets often lack fine-grained trajectory information and frame-level descriptions of driving maneuvers \cite{arai2025covla}, and with degraded inputs, VLMs can generate plausible but ungrounded responses \cite{xie2025drivebench}. Crucially, no existing work systematically studies how input configuration factors impact VLM performance on sequential driving scenes \cite{narayanan2019dynamic, atakishiyev2024explainable}.\\
\begin{figure}[!t]
    \centering
    \includegraphics[width=0.90\linewidth]{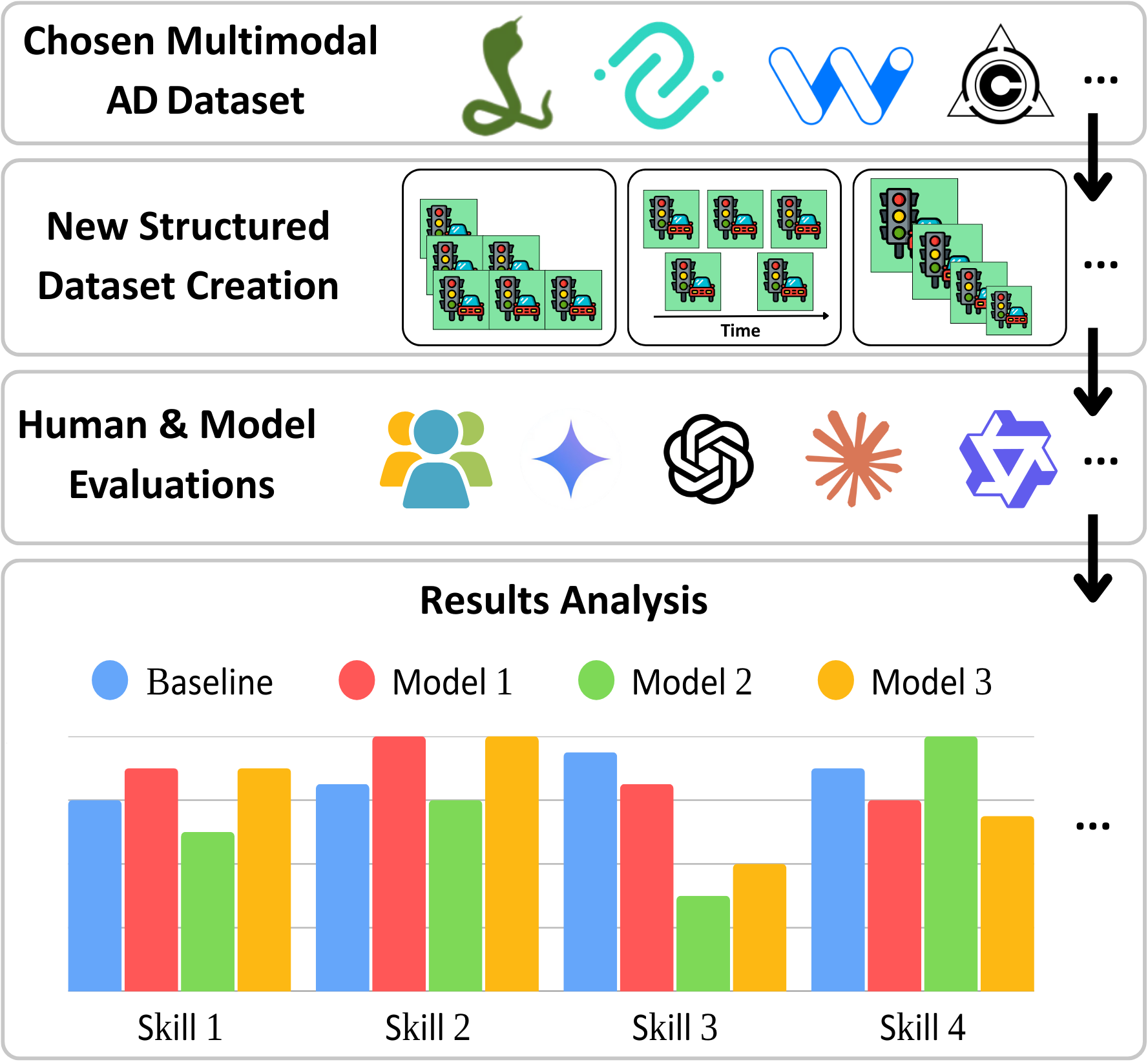}
    \vspace{-0.5em}
    \caption{VENUSS framework overview. Starting from driving datasets, VENUSS generates structured evaluation data with controlled variations in image count, timing, resolution, layout, and presentation mode. It evaluates both VLMs and humans on identical tasks across custom categories, identifies optimal input configurations per model, and establishes performance baselines.}
    \label{fig:venuss_overview}
    \vspace{-1em}
\end{figure}
To address this gap, we introduce VENUSS, a framework for systematic sensitivity analysis of VLM performance on sequential driving scenes. Our \textbf{key contributions} are:\\
(1) Sensitivity analysis: systematic evaluation of how image resolution, temporal sampling, frame count, spatial layout, and presentation mode affect VLM performance on sequential driving scenes.\\
(2) VENUSS framework: a dataset-agnostic evaluation pipeline supporting multiple driving datasets (CoVLA \cite{arai2025covla}, Honda Scenes \cite{narayanan2019dynamic}, NuScenes \cite{caesar2020nuscenes}, Waymo Open Dataset \cite{sun2020waymo}) with an extensible architecture for new datasets through minimal code modification.\\
(3) Human baselines: evaluation of humans on identical tasks and conditions as VLMs, via a configurable web application that also serves as a data curation tool.\\
(4) Large-scale benchmarking: evaluation of 25+ VLMs across 2,600+ scenarios, revealing that top models achieve only 57\% accuracy on basic perception tasks.\\
(5) Public code release: full pipeline with dataset generation, evaluation tools, and annotations for reproducibility.



\section{Related Work}
\label{sec:related_work}

This section reviews recent advancements in VLMs for autonomous driving, relevant datasets and benchmarks, and human perception in evaluation.\\
\textbf{VLMs in Autonomous Driving.} In recent years, VLMs have been increasingly studied for AD. The Vision Language Planning (VLP) framework \cite{pan2024vlp} introduces language models into vision-based motion planning, showing significant reductions in collision rates. Comprehensive reviews underline the VLMs' potential across the entire AD stack \cite{li2025applications, cui2024survey}, including perception enhancement with LiDAR data, trajectory prediction as language modeling, and decision-making with natural language explanations. Specific architectures like VERDI \cite{feng2025verdi} distill VLM reasoning into AD stacks to overcome computational demands, while AlphaDrive \cite{jiang2025alphadrive} employs reinforcement learning for enhanced planning. Efficiency-focused approaches include EM-VLM4AD for lightweight question answering \cite{emvlm4ad2024}, ReasonDrive for explicit reasoning \cite{chahe2025reasondrive}, and knowledge distillation techniques \cite{cao2025movekd}. Broader applications extend to robotic navigation, with the papers in the VLMNM 2024 workshop \cite{dipalo2024language, honerkamp2024language, hu2024deploying} and NaVILA \cite{cheng2025navila}, demonstrating VLMs' versatility in translating linguistic commands into actionable plans.\\
\textbf{Dynamic Scene Understanding: Datasets \& Benchmarks.} Robust VLM development depends on comprehensive datasets and evaluation benchmarks. The CoVLA dataset \cite{arai2025covla} provides 10,000 video clips with frame-level captions and trajectory actions, but lacks fine-grained trajectory information for vehicle maneuvers. The Honda Scenes Dataset \cite{narayanan2019dynamic} offers 80 hours of annotated driving videos for dynamic scene retrieval using CLIP models. Other relevant datasets include MARS for multi-agent interactions \cite{li2024multiagent}, LaMPilot-Bench for language model programs \cite{ma2024lampilot}.
VLADBench \cite{li2025finegrained} introduces fine-grained evaluation through hierarchically structured Question Answering (QA) tasks, from static knowledge to dynamic reasoning. While VLADBench addresses fine-grained assessment needs, its scope does not include different input image configurations (varying resolutions, temporal intervals, grid formats), which can significantly impact VLM performance in dynamic driving scenarios. DriveBench \cite{xie2025drivebench} evaluates the VLMs' reliability across clean, corrupted, and text-only inputs. It reveals how VLMs may generate plausible but ungrounded responses under degraded conditions. Our VENUSS complements such studies by analyzing the VLMs' performance across input variations, rather than corruption.\\
\textbf{Spatio-Temporal Reasoning.} Vehicle navigation requires spatio-temporal reasoning. ST-VLM \cite{ko2025st} addresses the \textit{temporal gap} in VLMs' evaluation by introducing STKit and STKit-Bench for kinematic instruction tuning, with 3D motion annotations including distance, speed, and direction. While ST-VLM highlights the importance of temporal reasoning, our VENUSS integrates temporal variations (time intervals between images, number of images) as part of a broader analysis of the VLM performance. TG-LLM \cite{xiong2024large} proposes temporal graph representations for sequential reasoning. These advancements highlight the need for open frameworks to assess the VLMs' ability to interpret temporal changes (such as acceleration, deceleration, and directional shifts), which existing benchmarks cannot fully evaluate.\\
\textbf{Human-in-the-Loop (HITL) and Human Perception.} Human expertise integration is crucial for AD development, particularly for safety and trustworthiness \cite{kumar2024applications, retzlaff2024human}. HITL methods include active learning for optimized data annotation \cite{settles2009active}, HITL reinforcement learning for real-time guidance \cite{retzlaff2024human}, and explainable AI for decision-making \cite{atakishiyev2024explainable}. Despite challenges including human bias, limited scalability, and integration complexity, HITL systems offer opportunities for improved accuracy and accountability. Human performance baselines provide references for VLM evaluation, particularly for nuanced interpretation of dynamic scenarios and temporal cues, a key component of our VENUSS framework.
\begin{figure*}[]
    \centering
    \includegraphics[width=\linewidth]{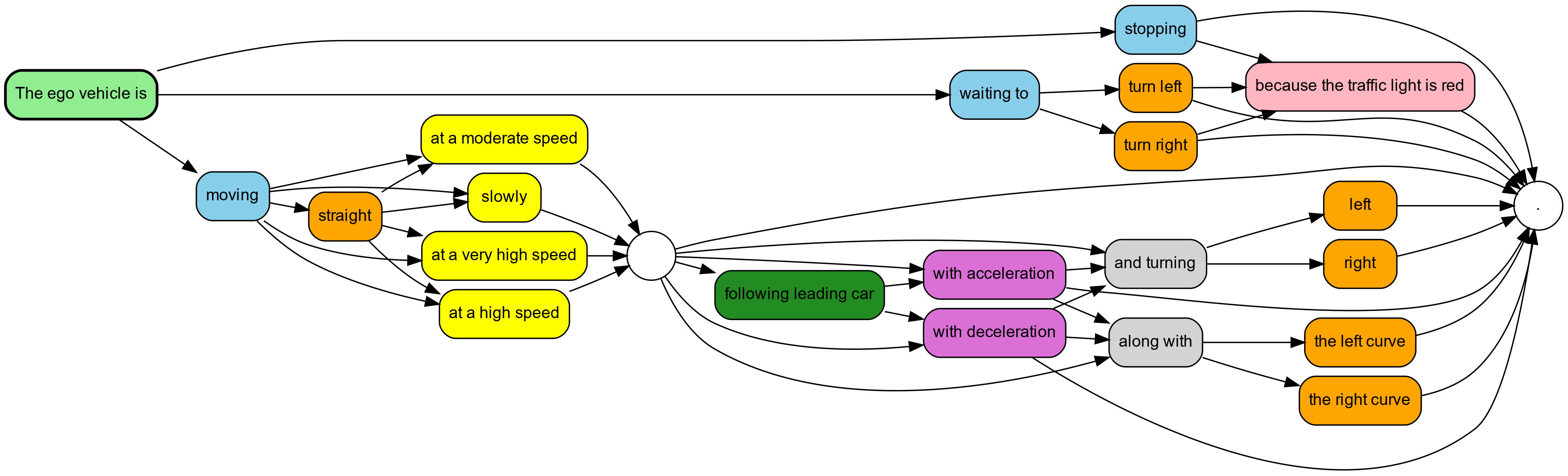}
    \caption{Visualization of all CoVLA textual descriptions automatically categorized by VENUSS. The framework identified seven categories from the natural language descriptions: motion types (blue), velocity descriptors (yellow), directional behaviors (orange), acceleration patterns (purple), following behavior (dark green), traffic light conditions (light red), and road curvature detection (gray).}
    \label{fig:state_transition}
\end{figure*}
\section{Methodology}
\label{sec:methodology}


\subsection{VENUSS Framework Overview}
\label{subsec:overview}

VENUSS consists of four modules (Fig.~\ref{fig:venuss_overview}): dataset generation pipeline, fine-grained annotation system, human baseline establishment, and prompt design. Starting from driving videos with textual descriptions, we extract temporal frame sequences and transform them into structured image collages for systematic VLM evaluation.

\subsection{Dataset Generation Pipeline}
\label{subsec:component1}

This module extracts sequential frames from videos of driving scenarios at controlled temporal intervals, and outputs structured image collages that capture temporal relations. Specifically, our pipeline samples video frames at specified intervals (100 ms to 1000 ms apart), creating discrete temporal sequences representing the evolution of driving scenarios over time. 
%
In a compact form:
\begin{equation}
    \mathbf{C} = \text{Generate}(\mathbf{V}, \mathbf{T}, \mathbf{N}, \mathbf{R}, \mathbf{G})
    \label{eq:component1}
\end{equation}
where $\mathbf{C}$ is the generated image collage, $\mathbf{V}$ is the input video from which frames are extracted, $\mathbf{T}$ is the temporal sampling of the extracted frames (10 possible levels: from 100 to 1000 ms), $\mathbf{N}$ is the frame count (from 1 to 10 frames), $\mathbf{R}$ is the image resolution (160×90, 320×180, 480×270, 640×360, 960×540, 1920×1080), and $\mathbf{G}$ is the grid spatial arrangement, which tests all possible $m \times n$ configurations where $m \times n \leq 10$ (1×1, 1×2, 1×3, \dots{}, 1×10, 2×1, 2×2, \dots{}, 2×5, 3×1, 3×2, 3×3, 4×1, 4×2, 5×1, 5×2, 6×1, \dots{}, 10×1). 

\subsection{Fine-Grained Annotation System}
\label{subsec:component2}

The second module transforms textual descriptions from driving datasets into structured, fine-grained labels for VLM evaluation. VENUSS is designed to be dataset-agnostic, requiring minimal adaptation for new datasets through modular configuration of annotation categories and semantic mapping functions.

The categories are dataset-specific and dependent on the content of the dataset's textual descriptions by default. However, they are also user-configurable, enabling adaptation to different annotation schemes and evaluation objectives.

We release VENUSS with the configuration files necessary to run the framework on CoVLA \cite{arai2025covla} and Honda Scenes \cite{narayanan2019dynamic}, with extensibility for additional datasets through a three-file modification process (dataset configuration, annotation parser, and evaluation questions). 

Our annotation process transforms the natural language descriptions into structured evaluation labels:
\begin{equation}
    \mathbf{A} = \text{Extract}(\mathbf{D}, \mathbf{Q}, \mathbf{M})
    \label{eq:component2}
\end{equation}
where $\mathbf{A}$ are the structured annotations, $\mathbf{D}$ are the original textual descriptions, $\mathbf{Q}$ are the categories, and $\mathbf{M}$ implements the semantic mapping function that translates natural language phrases to categorical answer options. These structured annotations serve as ground truth labels for systematic VLM evaluation, as detailed in the subsequent sections. For example, the caption ``The ego vehicle is moving straight at a high speed'' is mapped to the answer key ``AABACBB'': moving (A), straight (A), high speed (B), not following (A), no acceleration (C), no traffic light (B), no curve (B), as illustrated in Figure~\ref{evaluation_interface}.

In this paper, we demonstrate the framework using CoVLA dataset results: the framework classifies the textual descriptions into seven categories capturing ego vehicle state and driving scenarios: (1) motion state (moving, stopping, stopped); (2) directional motion (moving straight, left, right); (3) velocity (very high to low); (4) following behavior (is the ego vehicle following another agent?); (5) acceleration (positive, negative or zero); (6) detection of traffic lights; (7) ego vehicle's motion along a curved or straight road.

These seven categories were automatically extracted from the first sentence of CoVLA's captions and align with fundamental aspects of autonomous driving evaluation: basic kinematic understanding \cite{ko2025st}, trajectory assessment \cite{arai2025covla}, speed perception \cite{li2025applications}, multi-agent interaction \cite{cui2024survey}, dynamic state changes \cite{ko2025st}, environmental awareness \cite{atakishiyev2024explainable}, and path understanding \cite{narayanan2019dynamic}. This scope was deliberately chosen for presentation purposes; using the full CoVLA caption text would automatically produce several additional categories. The framework also generalizes across annotation schemes, Honda Scenes \cite{narayanan2019dynamic} for instance yields 16 environmental categories (road type, weather, surface, lighting, infrastructure), complementing CoVLA's behavioral categories. Sample evaluations for each dataset are available on the supplementary webpage.

Figure~\ref{fig:state_transition} demonstrates the framework's automatic categorization of CoVLA's textual descriptions, identifying 108 distinct driving scenarios across these seven categories. The deterministic keyword-based mapping was manually curated, with all 108 scenarios verified against the original driving videos using the web application described in Section~\ref{subsec:human_evaluation}.

\subsection{Human Baseline Establishment}
\label{subsec:human_evaluation}

The third module establishes human performance baselines through a configurable web application that adapts to the dataset being used and its evaluation scenarios. The interface presents sequential driving scenarios followed by multiple-choice questions corresponding to the annotation categories defined in Section \ref{subsec:component2}. Beyond baseline collection, the interface serves as a versatile tool for data curation and annotation validation.

\begin{figure}[!t]
    \centering
    \includegraphics[width=\linewidth]{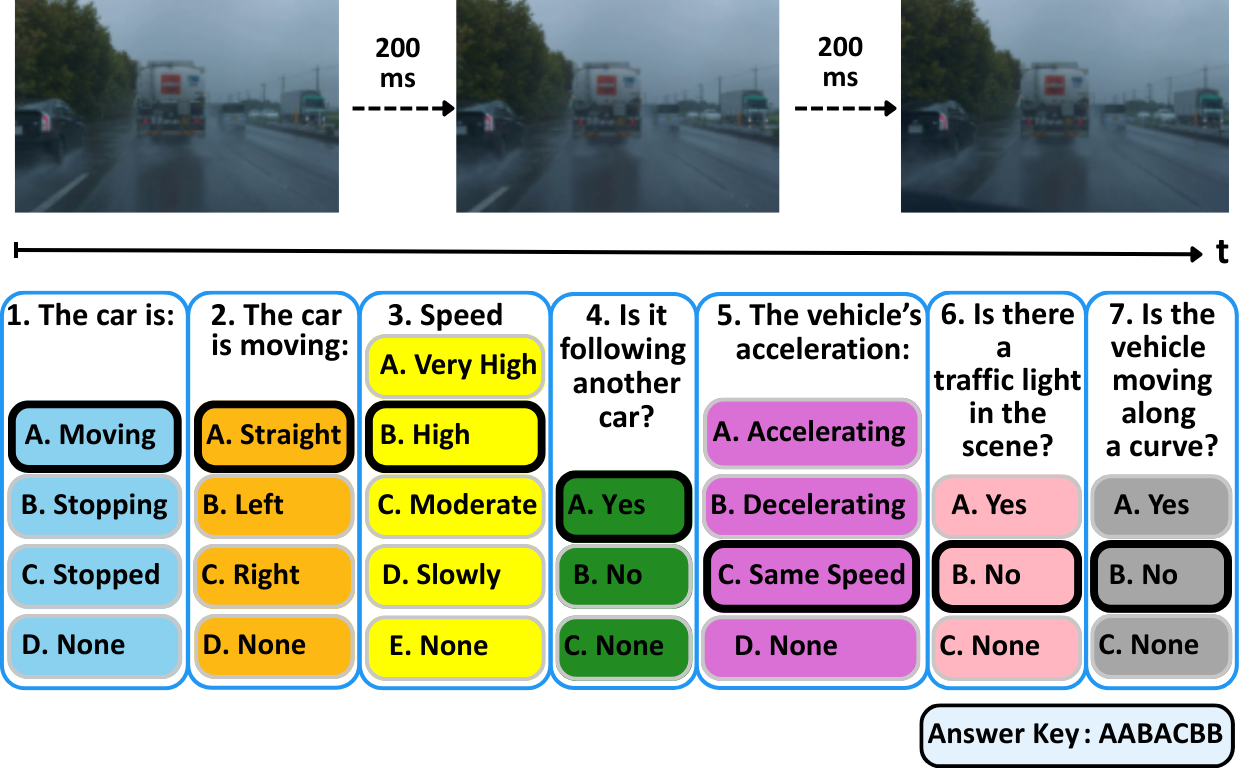}
    \caption{Simplified example of the human evaluation questionnaire with the seven-category format (Sec. \ref{subsec:component2}). The interface presents sequential driving images with temporal intervals, and seven multiple-choice questions for the corresponding seven categories. The final answer key (bottom right) concatenates the responses for comparison with ground truth annotations.}
    \label{evaluation_interface}
\end{figure}

The web application supports three presentation modes: (1) image collages matching the VLM input format for direct performance comparison, (2) animated GIFs providing temporal continuity and more intuitive scenario understanding for human evaluators, and (3) video playback for annotation verification and data curation. As shown in Figure~\ref{evaluation_interface}, participants view driving scenarios and answer structured questions across all annotation categories. Responses are concatenated into answer keys (e.g., "AABACBB") for direct comparison with ground truth annotations and VLM outputs.

The modular design enables adaptation to different datasets by modifying question sets, image sources, and evaluation protocols. Beyond performance evaluation, the web interface facilitates data quality assessment, annotation verification, and identification of ambiguous scenarios by human experts.

To demonstrate the framework's application, we conducted human evaluations using both presentation modes. Initially, three participants with diverse backgrounds (researcher in autonomous systems, computer vision graduate student, and automotive engineer) completed evaluations using image collages, following the same format used for VLM evaluation. Subsequently, five additional participants with similar backgrounds and valid driving licenses (3-8 years experience) completed 108 evaluations each using the GIF interface, covering all configuration combinations (frame counts, temporal intervals, resolutions, and spatial layouts).

\subsection{VLM Prompt Design}
\label{subsec:prompt_design}

The fourth module implements our prompt design, which provides VLMs with contextual information while maintaining standardized evaluation conditions.

\textbf{System Prompt:} Our system prompt establishes the evaluation context: "You are an expert in autonomous driving scenario analysis. You will be shown images representing sequential driving scenarios and must classify them according to specific categories."

\textbf{User Prompt:} The user prompt provides context about the image content and task requirements. For each evaluation, models receive: (1) Description of the image configuration (e.g., "You are viewing a 2×3 grid of images showing a driving scenario captured at 200 ms intervals"), (2) Temporal context ("The images are arranged chronologically from left to right, top to bottom"), (3) The classification questions corresponding to the dataset's annotation categories (Section \ref{subsec:component2}), each with their answer options, and (4) formatting instructions requesting that models include a short answer key at the end of their response in the format: "1) [letter] 2) [letter] ... 7) [letter]". Models are free to provide full explanations; only the answer key is parsed for evaluation.

\textbf{Image Context:} Models are informed about the image source ("These images are from the ego vehicle's perspective in a driving scenario") and the configuration details (grid layout, temporal spacing, resolution level), enabling them to understand the spatial-temporal relations among frames.

Our prompt design ensures all models receive identical information and instructions, enabling fair comparison while providing sufficient context for scenario interpretation. Since the evaluated tasks are inherently categorical (e.g., is the vehicle moving, stopping, or stopped?), extracting discrete answers enables reproducible and scalable evaluation across 25+ models and 2,600+ scenarios. Evaluating full free-form responses, e.g. via LLM-as-judge, is left as future work. Full prompt templates for each dataset are available on the supplementary webpage.





\section{Experiments}
\label{sec:experiments}

\begin{table*}[!t]
\centering
\caption{Five-phase experimental protocol for systematic VLM evaluation.}
\label{tab:experimental_phases}
\begin{tabular}{@{}llll@{}}
\toprule
\textbf{Phase} & \textbf{Target Variable} & \textbf{Varied Parameters} & \textbf{Fixed Parameters} \\
\midrule
1 & Resolution & 6 resolution levels & 2×2 grid layout, 200ms intervals, collage mode \\
2 & Frame Count & 1-10 frames & Level 1 \& Level 6 resolution, 200ms intervals, 1×N layout \\
3 & Temporal & 100-1000ms intervals & Level 1 \& Level 6 resolution, 4 frames, 1×4 layout \\
4 & Spatial & All possible grid layouts & Level 1 \& Level 6 resolution, 200ms intervals \\
5 & Presentation & Collage, separate, batch modes & Level 1 \& Level 6 resolution, 2×2 grid, 200ms intervals \\
\bottomrule
\end{tabular}
\end{table*}




\subsection{Research Questions and Evaluation} \label{subsec:research_questions}

Our framework assesses VLM performance across different configurations and compares models in varying conditions.
We address five fundamental research questions (RQ) about VLM performance in dynamic driving scenarios:\\
\textbf{RQ1:} How does image resolution impact VLM understanding of driving scenarios?\\
\textbf{RQ2:} What is the optimal number of sequential frames for temporal comprehension?\\
\textbf{RQ3:} How do temporal intervals between frames affect scenario interpretation?\\
\textbf{RQ4:} Which spatial arrangement of multiple frames maximizes VLM performance?\\
\textbf{RQ5:} How do different presentation modes influence evaluation outcomes?

To answer these questions, we vary five key dimensions: \textit{Resolution} (6 levels: 160×90 to 1920×1080), \textit{Frame Count} (1-10 frames), \textit{Temporal Intervals} (100 ms to 1000 ms in 100ms increments), \textit{Grid Layouts} (all feasible arrangements for N frames: 1×N, N×1, and rectangular grids), and \textit{Presentation Modes} (collage, separate images, batch processing).

\subsubsection{Experimental Design and Phase Structure}

We use a five-phase protocol to progressively explore the configuration space, as detailed in Table~\ref{tab:experimental_phases}.

Each phase runs 10 evaluations per configuration, to balance statistical reliability and computational efficiency. The frame count is bounded at 10 as a controlled experimental choice; as shown in Section~\ref{sec:results}, VLM performance plateaus at 3-4 frames, indicating that this cap does not limit the analysis. Human evaluators are tested under the same conditions, ensuring a fair comparison. This structured analysis reveals how configuration choices significantly affect performance, underscoring the need for systematic assessment.

\subsection{Evaluation Metrics and Assessment}
\label{subsec:component3}

Our evaluation uses a flexible composite metric that can be customized based on application requirements, in the form of a weighted sum of all classification categories:
\begin{align}
    \text{Score} &= \alpha_1 \, S_1 + \alpha_2 \, S_2 + \dots + \alpha_7 \, S_7 = \sum_{i=1}^{7} \alpha_i \, S_i&
    \label{eq:loss}
\end{align}
where $S_i$ is the accuracy score in each classification category (Section \ref{subsec:component2}): $S_1$ (motion state), $S_2$ (direction), $S_3$ (speed), $S_4$ (following behavior), $S_5$ (acceleration), $S_6$ (traffic light detection), and $S_7$ (curve navigation). The weighting factors $\{\alpha_1,\dots,\alpha_7\}$ can be tuned to the specific application. For our evaluation, we employ equal weighting ($\alpha_i = 1/7, \forall i$) to provide unbiased model comparison.

During inference, VLMs are presented with image collages generated by our evaluation framework, and asked to classify each of the seven categories based on their understanding of the sequential driving scenario. The model performance is measured against both ground truth annotations and human baselines.

\section{Results and Discussion}
\label{sec:results}

This section presents the results of our systematic evaluation with 25+ VLMs across 2,600+ scenarios of the CoVLA dataset \cite{arai2025covla}. We analyze performance variations across multiple dimensions and compare VLM capabilities with human baselines, revealing critical insights about input configurations and their impact on VLM performance.

\subsection{Overall Performance Analysis}
\label{subsec:quantitative}

Table~\ref{tab:comprehensive_model_performance} reports the results for all the evaluated VLMs, in terms of overall accuracy, precision, recall and F1 score on the seven selected categories of Section~\ref{subsec:component2}, F1 scores for each category, and average query times. Note that the values in the table are averaged over all the tested configurations of the input image sequences: the sensitivity to the input configurations is analyzed in Section~\ref{subsec:configuration_analysis}.

The results reveal that top-performing VLMs achieve competitive performance compared to human baselines on these basic perception tasks. Qwen-VL-Max achieves the best result, with 57\% accuracy on our benchmark compared to the human baseline of 62.5\%. Among human evaluators, the GIF-based evaluations consistently outperformed collage-based evaluations (65\% vs 56\% peak accuracy), with this 16\% relative improvement confirming that presentation format significantly impacts human performance. The modest gap between human and model performance demonstrates that while intuitive presentation formats can enhance human results, the underlying challenges of reasoning about complex driving scenarios remain substantial across both humans and VLMs.

\begin{table*}[!t]
\centering
\caption{VLMs performance evaluation, with the unified metric score of eq. \eqref{eq:loss} (columns 2-5), F1 scores for our seven categories (motion state, direction, speed, following behavior, acceleration, traffic light detection, road curvature: columns 6-12), and query times.}
\label{tab:comprehensive_model_performance}
\small
\begin{tabular}{@{}l|cccc|ccccccc|c@{}}
\toprule
\textbf{Model / Human} & \multicolumn{4}{c|}{\textbf{Overall Metrics}} & \multicolumn{7}{c|}{\textbf{F1 score for each category}} & \textbf{Avg Query} \\
 & \textbf{Acc.} & \textbf{Prec.} & \textbf{Rec.} & \textbf{F1} & \textbf{Mot.} & \textbf{Dir.} & \textbf{Spd.} & \textbf{Fol.} & \textbf{Accel.} & \textbf{TL} & \textbf{Crv.} & \textbf{Time (s)} \\
\midrule
evaluator-1 (collage) & \underline{0.63} & \underline{0.66} & \underline{0.63} & \underline{0.64} & 0.88 & \underline{0.56} & 0.32 & 0.62 & \textbf{0.40} & \textbf{0.89} & 0.69 & N/A \\
evaluator-2 (collage) & 0.57 & 0.61 & 0.57 & 0.58 & 0.87 & 0.50 & \underline{0.41} & 0.64 & 0.37 & 0.76 & 0.40 & N/A \\
evaluator-3 (collage) & 0.54 & 0.59 & 0.54 & 0.55 & 0.82 & 0.27 & 0.22 & 0.60 & 0.27 & 0.76 & 0.68 & N/A \\
evaluator-4 (gif) & \textbf{0.65} & \textbf{0.68} & \textbf{0.65} & \textbf{0.66} & \underline{0.92} & 0.42 & \textbf{0.49} & \textbf{0.80} & 0.27 & 0.85 & \textbf{0.74} & N/A \\
evaluator-5 (gif) & 0.63 & 0.66 & 0.63 & 0.64 & 0.88 & 0.56 & 0.32 & 0.62 & 0.40 & \underline{0.89} & 0.69 & N/A \\
evaluator-6 (gif) & 0.63 & 0.66 & 0.63 & 0.64 & \textbf{0.93} & \textbf{0.58} & 0.37 & \underline{0.75} & 0.30 & 0.79 & \underline{0.72} & N/A \\
evaluator-7 (gif) & 0.63 & 0.66 & 0.63 & 0.63 & 0.91 & 0.51 & \underline{0.45} & 0.72 & \underline{0.32} & 0.83 & 0.70 & N/A \\
evaluator-8 (gif) & 0.62 & 0.64 & 0.62 & 0.62 & 0.91 & 0.54 & 0.42 & 0.72 & 0.31 & 0.84 & 0.62 & N/A \\
\midrule[1pt]
qwen-vl-max & \textbf{0.57} & \underline{0.62} & \textbf{0.57} & \textbf{0.59} & \underline{0.90} & 0.14 & \underline{0.31} & 0.60 & 0.12 & 0.88 & 0.68 & 2.4 \\
claude-3.7-sonnet-latest & 0.55 & 0.60 & 0.55 & \underline{0.57} & 0.88 & 0.17 & 0.20 & 0.59 & 0.10 & 0.89 & 0.64 & 2.8 \\
claude-3.5-sonnet-latest & 0.55 & 0.60 & 0.55 & 0.57 & 0.88 & 0.19 & 0.17 & 0.60 & 0.09 & 0.90 & 0.66 & 2.6 \\
gpt-4o-mini & \underline{0.55} & 0.60 & \underline{0.55} & 0.56 & 0.85 & 0.22 & 0.15 & \textbf{0.63} & \textbf{0.25} & 0.87 & 0.68 & 3.0 \\
qwen-vl-plus & 0.54 & 0.61 & 0.54 & 0.56 & 0.85 & \textbf{0.33} & 0.12 & 0.48 & 0.08 & 0.90 & 0.63 & \textbf{1.6} \\
gemini-1.5-flash & 0.54 & 0.60 & 0.54 & 0.56 & 0.87 & 0.22 & 0.14 & 0.45 & 0.11 & 0.88 & \underline{0.71} & 2.6 \\
qwen2.5-vl-7b-instruct & 0.54 & \textbf{0.62} & 0.54 & 0.56 & 0.82 & 0.28 & 0.15 & 0.47 & 0.09 & 0.88 & 0.64 & 2.2 \\
claude-3-opus & 0.54 & 0.59 & 0.54 & 0.55 & \textbf{0.91} & 0.08 & 0.21 & 0.57 & 0.08 & \textbf{0.90} & 0.50 & 4.1 \\
qwen2.5-vl-72b-instruct & 0.52 & 0.59 & 0.52 & 0.55 & 0.88 & 0.17 & 0.23 & 0.59 & 0.10 & 0.77 & 0.67 & 3.7 \\
gemini-2.0-flash-lite & 0.53 & 0.59 & 0.53 & 0.54 & 0.81 & 0.26 & 0.17 & 0.57 & 0.15 & 0.84 & 0.65 & 2.8 \\
claude-opus-4-0 & 0.52 & 0.59 & 0.52 & 0.54 & 0.84 & 0.15 & 0.19 & 0.52 & 0.09 & 0.89 & 0.60 & 7.1 \\
gemini-2.0-flash-exp & 0.51 & 0.61 & 0.51 & 0.53 & 0.88 & 0.26 & 0.13 & \underline{0.60} & 0.12 & 0.72 & 0.68 & 4.3 \\
gemini-1.5-pro & 0.50 & 0.60 & 0.50 & 0.53 & 0.83 & 0.17 & 0.16 & 0.56 & 0.14 & 0.84 & 0.65 & 4.1 \\
claude-sonnet-4-0 & 0.51 & 0.55 & 0.51 & 0.53 & 0.85 & 0.11 & 0.15 & 0.52 & 0.08 & 0.88 & 0.59 & 6.1 \\
gemini-2.0-flash & 0.49 & 0.61 & 0.49 & 0.51 & 0.88 & 0.25 & 0.13 & 0.59 & 0.13 & 0.62 & 0.68 & 2.8 \\
qwen2.5-vl-32b-instruct & 0.48 & 0.58 & 0.48 & 0.51 & 0.82 & \underline{0.29} & \textbf{0.32} & 0.45 & 0.11 & 0.80 & 0.61 & 4.4 \\
qwen2.5-vl-3b-instruct & 0.51 & 0.54 & 0.51 & 0.51 & 0.87 & 0.21 & 0.16 & 0.29 & 0.07 & \underline{0.90} & 0.53 & \underline{1.9} \\
gemini-1.5-flash-8b & 0.46 & 0.58 & 0.46 & 0.50 & 0.85 & 0.14 & 0.13 & 0.51 & \underline{0.16} & 0.61 & \textbf{0.73} & 2.5 \\
claude-3-sonnet & 0.45 & 0.53 & 0.45 & 0.48 & 0.73 & 0.11 & 0.13 & 0.38 & 0.08 & 0.80 & 0.59 & 2.6 \\
gpt-4o & 0.42 & 0.60 & 0.42 & 0.47 & 0.73 & 0.14 & 0.12 & 0.54 & 0.08 & 0.75 & 0.63 & 3.0 \\
claude-3.5-haiku-latest & 0.42 & 0.55 & 0.42 & 0.46 & 0.78 & 0.15 & 0.20 & 0.49 & 0.10 & 0.77 & 0.59 & 3.1 \\
\bottomrule
\end{tabular}
\end{table*}
A statistically significant performance hierarchy among model families is observed: Qwen (52.8\%) $>$ Gemini (51.2\%) $>$ Claude (50.5\%) $>$ GPT (49.4\%). Qwen architectures demonstrate both superior accuracy and consistency ($\sigma$=0.032) compared to GPT models ($\sigma$=0.082), indicating 2.56$\times$ better reliability. However, substantial performance variability persists across VLMs ($\sigma$=0.212 overall), suggesting that careful validation is required before practical use.

Notably, GPT-4o-Mini significantly outperforms the flagship GPT-4o by 13 percentage points (55\% vs 42\%), challenging assumptions about model size correlating with performance on spatial-temporal understanding tasks. This counterintuitive result is explained by GPT-4o's frequent refusals, where it consistently selected the ``do not apply / none of the above'' option, lowering F1 scores. Similarly, some advanced models such as Gemini-2.5-Pro exhibited excessive non-responsiveness (over 90\% refusals), necessitating exclusion from the study. Although modified prompts were tested, they were rejected to maintain identical evaluation conditions across all models. These behaviors suggest that stronger safety filters in larger models can interfere with task compliance, while smaller models provide more consistent responses. This is itself a relevant finding: safety guardrails in some proprietary models make them unsuitable for driving-related evaluation, regardless of their actual capability.

Category-wise, both VLMs and humans show F1 scores below 70\% for most tasks, except for motion state (0.73–0.91) and traffic light detection (0.61–0.90). In contrast, temporal reasoning tasks remain highly challenging: acceleration detection (0.07–0.25) and direction estimation (0.08–0.33) show consistently low performance. The similar performance levels between humans and the best-performing VLMs indicate genuine reasoning challenges in driving scenarios rather than dataset-specific biases.

The performance gap between VLMs and humans remains at 18.2\%, indicating that current models have not yet reached human-level performance on these tasks. Nevertheless, our configuration analysis reveals a 48.2\% improvement potential through optimized input settings, suggesting that proper configuration of input parameters can significantly improve VLM performance without requiring fundamental architectural changes.

\begin{figure*}[!t]
\centering
\includegraphics[width=\textwidth]{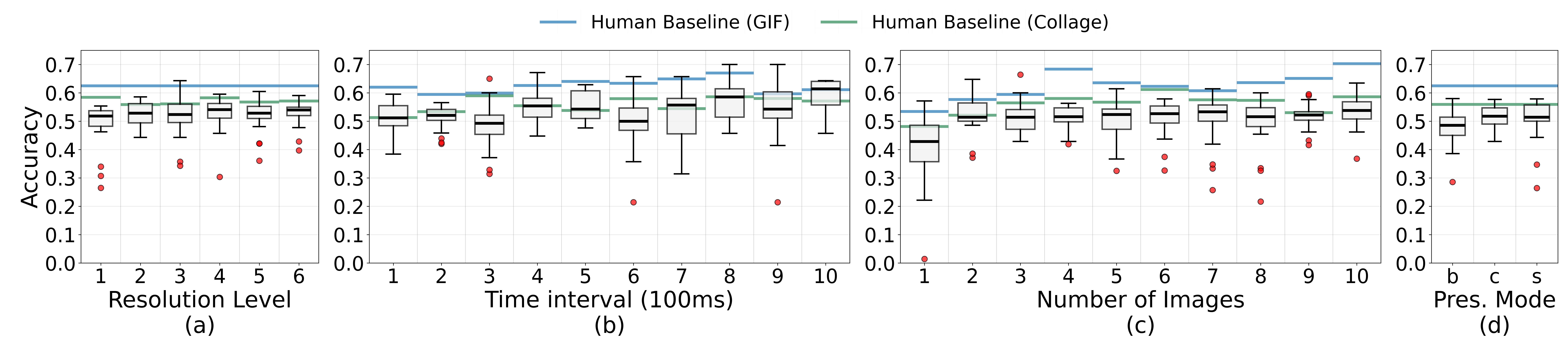}
\caption{Performance analysis across 4 dimensions as described in \ref{subsec:research_questions}. (a) Performance by resolution level (1-6), (b) Performance by time interval (100ms increments, 1-10), (c) Performance by number of images (1-10), and (d) Performance by presentation mode (b=batch, c=collage, s=separate). Horizontal lines show human baselines: green for collage-based evaluation, blue for GIF-based evaluation. Box-and-whisker plots summarize the performance distribution of all 25+ evaluated VLMs for each configuration, with red dots indicating outliers.}
\label{fig:comprehensive_performance}
\end{figure*}

Figure~\ref{fig:comprehensive_performance} provides a comprehensive overview of model performance across all experimental conditions. The figure shows VLM performance compared to human baselines across (a) resolution levels, (b) time intervals, (c) number of images, and (d) presentation modes. The horizontal lines indicate human performance baselines for both collage-based (green) and GIF-based (blue) evaluations, while the box plots show the distribution of VLM performance. Red dots represent statistical outliers in the VLM performance distributions.

\subsection{Configuration Analysis} \label{subsec:configuration_analysis}

Our systematic evaluation across 270 unique configuration combinations reveals that parameter optimization significantly impacts VLM performance, with statistical analysis showing a 48.2\% relative improvement potential through systematic configuration tuning. This section analyzes how each configuration dimension affects performance.

\begin{figure}[]
    \centering
    \includegraphics[width=\linewidth]{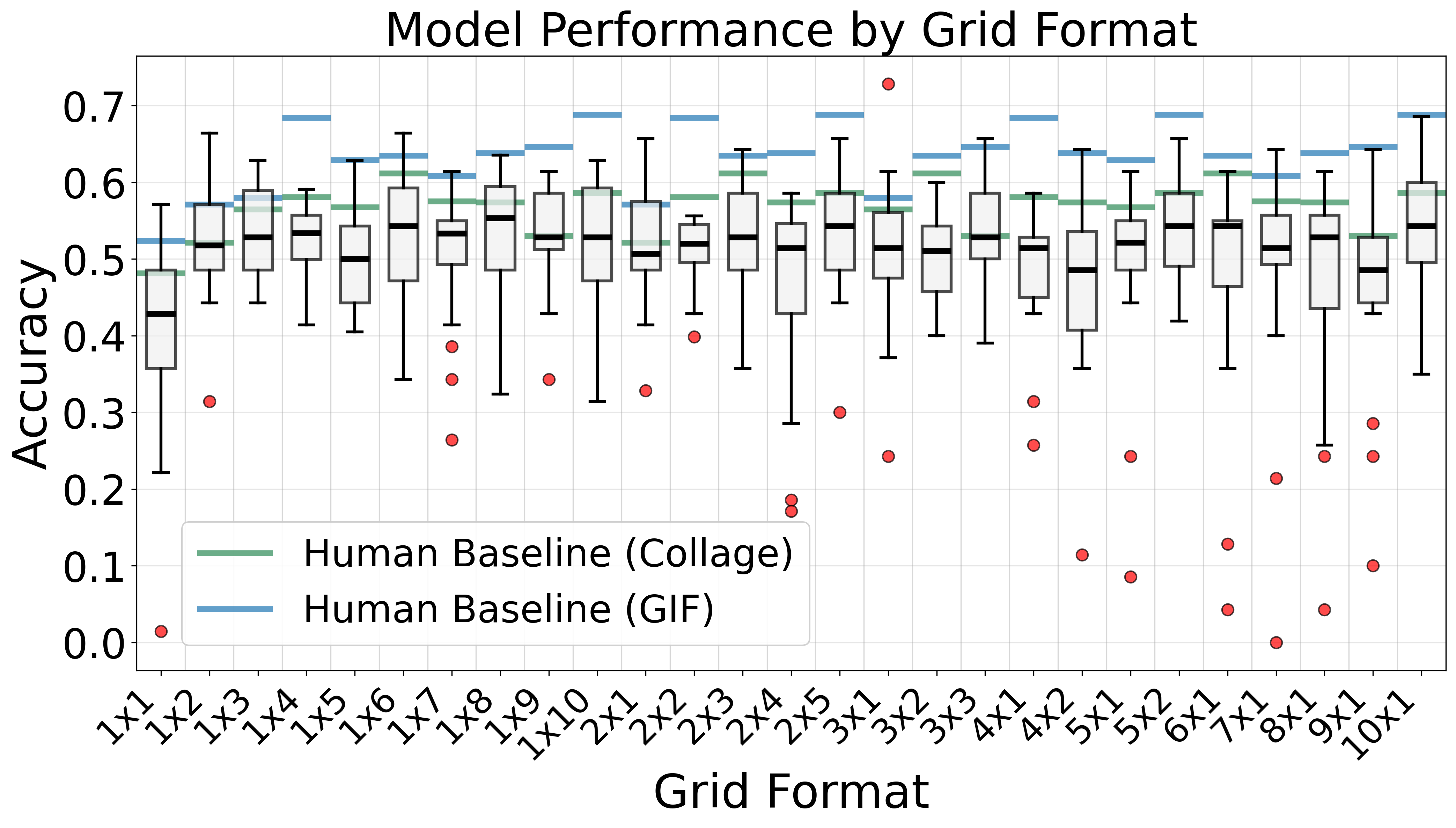}
    \caption{Performance comparison across different grid formats for the image sequences. Specific grid layouts demonstrate better performance for temporal understanding tasks.}
    \label{fig:grid_format}
\end{figure}


Resolution analysis (Figure~\ref{fig:comprehensive_performance}a) shows 720p provides optimal balance, achieving 95\% of peak performance while maintaining computational tractability. Lower resolutions likely lose critical visual details needed for scene interpretation, while higher resolutions add pixel density without new semantic information for these tasks. Qwen architectures demonstrate superior robustness across resolutions ($\sigma$=0.032) versus GPT models ($\sigma$=0.082).

Optimal performance is achieved at 1000ms intervals (0.5918 accuracy) versus 200ms intervals (0.5068 accuracy), indicating a 14.4\% accuracy penalty for shorter intervals (Figure~\ref{fig:comprehensive_performance}b). Longer intervals capture more visually distinct frames, making scene changes more apparent, while short intervals produce near-duplicate frames that provide limited additional information.

Increasing the number of frames from 1 to 3-4 yields a 31.5\% improvement (Figure~\ref{fig:comprehensive_performance}c), but performance plateaus beyond 4 frames. This suggests current VLMs cannot effectively integrate information across long sequences, likely due to attention mechanisms that struggle to track temporal changes across many frames.

Collage presentation outperforms sequential processing by 6.0\% (Figure~\ref{fig:comprehensive_performance}d). Spatial co-location allows models to compare frames simultaneously within a single image, rather than relying on context memory across separate inputs.

Horizontal layouts (3$\times$1, 4$\times$1) outperform square grids (Figure~\ref{fig:grid_format}), as left-to-right arrangement matches natural temporal flow and avoids ambiguity in reading order that square grids introduce. All configuration dimensions show statistically significant performance differences (p < 0.001, $\eta^2$ = 0.41), confirming large, practically meaningful effect sizes.  

\subsection{Task-Specific Performance Analysis}

Our analysis shows notable performance changes across the seven evaluation categories (columns 6-12 in Table \ref{tab:comprehensive_model_performance}), revealing key strengths and weaknesses of current VLMs in driving scene understanding. We now examine these categories in increasing order of difficulty.\\
\textbf{Vehicle Motion Detection} (Easy): F1 scores range in 0.73-0.91, with most models above 0.80, showing that basic motion detection is a feasible task for current VLMs.\\
\textbf{Traffic Light Detection} (Moderate): F1 scores range in 0.61-0.90, with most models above 0.75, showing that static object detection is within the current VLMs' capability.\\
\textbf{Curved Road Detection} (Moderate-Hard): F1 scores range in 0.40-0.73, with most models below 0.70, showing moderate difficulties of VLMs in spatial reasoning.\\
\textbf{Car Following Behavior} (Hard): F1 scores range in 0.29-0.64, with most models below 0.60: understanding the relative motion of vehicles can be challenging for VLMs.\\
\textbf{Vehicle Direction Analysis} (Hard): F1 scores range in 0.08-0.33, with most models below 0.25, revealing major limitations in determining turning directions.\\
\textbf{Vehicle Speed Assessment} (Very Hard): F1 scores range in 0.12-0.32, with most models below 0.20: quantifying the temporal motion of vehicles is difficult for VLMs.\\
\textbf{Vehicle Acceleration Detection} (Hardest): F1 scores range in 0.07-0.25, with most models below 0.15, representing the most challenging task and confirming the limitations in understanding the vehicle dynamics.

\subsection{Practical Recommendations}

Based on our analysis, we provide practical recommendations for researchers and practitioners working with VLMs in driving-related tasks:

\subsubsection{Configuration Guidelines}

\textbf{Driver Assistance Systems (Non-Critical):} 720p resolution, 4-frame sequences, 300-500ms intervals using collage presentation. This configuration achieves 0.5510 $\pm$ 0.1758 accuracy while maintaining real-time viability ($\sim$2-3 frames per second (FPS)) for lane departure warnings or traffic monitoring.

\textbf{Offline Scene Analysis:} Optimal configuration (1000ms intervals, 4 frames, 960$\times$540 resolution) achieves 0.5918 accuracy for post-incident analysis, route planning validation, or training data annotation where real-time constraints are relaxed.

\textbf{Real-Time Supervisory Systems:} 14.4\% performance penalty with 200ms intervals to maintain 5 FPS processing for driver monitoring or secondary validation systems, achieving 0.5068 accuracy while preserving safety margins.

\subsubsection{Model Selection Criteria}

Our analysis provides quantitative guidance for AD system architects:

\textbf{High-Reliability Applications:} Qwen architectures demonstrate superior consistency ($\sigma$=0.032) and accuracy (52.8\%), particularly suitable for safety-adjacent functions requiring predictable performance.

\textbf{Cost-Constrained Deployments:} GPT-4o-Mini should be considered over flagship models, as our analysis reveals smaller models often outperform larger variants in spatial-temporal reasoning tasks while offering better computational efficiency.

\textbf{Multi-Modal Integration:} The 18.2\% human-VLM gap indicates VLMs should complement rather than replace traditional computer vision systems, with our configuration optimization providing the foundation for effective sensor fusion architectures.

\subsubsection{Validation Considerations}

The substantial performance variability ($\sigma$=0.212) necessitates rigorous validation protocols:

\textbf{Performance Bounds:} Accuracy ranges of 0.35-0.75 even in optimal configurations are expected, requiring fail-safe mechanisms for low-confidence scenarios.

\textbf{Scenario-Specific Validation:} Task hierarchy analysis reveals that acceleration detection (F1<0.25) and directional reasoning (F1<0.33) require additional validation or hybrid approaches for safety-critical applications.

\textbf{Configuration Robustness:} Systems must be deployed with adaptive configuration selection based on real-time performance monitoring, utilizing our identified optimal parameter spaces to maintain consistent operation across varying conditions.

\section{Conclusion}
\label{sec:conclusion}

This paper introduces VENUSS, an open-source, dataset-agnostic framework for systematic sensitivity analysis of VLM performance on sequential driving scenes. VENUSS generates structured evaluations by varying image count, temporal spacing, resolution, spatial layout, and presentation mode, and includes a web interface for human baseline collection and data curation. The framework and results are publicly released to support further research.

Evaluating 25+ VLMs across 2,600+ CoVLA scenarios reveals critical limitations: while VLMs achieve reasonable performance on static tasks (traffic light detection: F1 0.75--0.90), they struggle with temporal reasoning (acceleration detection: F1 0.07--0.25). Leading VLMs reach only 57\% accuracy versus human baselines of 54--65\%, with systematic parameter optimization yielding up to 48.2\% improvement potential. However, the 18.2\% gap below human baselines and high variability ($\sigma$=0.212) indicate current VLMs require careful validation before deployment.

This study evaluates general-purpose VLMs rather than AD-adapted models, so performance gaps may partly reflect domain mismatch; ground truth labels derive from CoVLA's captions rather than sensor-based annotations, introducing potential label noise; and the human baseline, while covering both collage and GIF modes, would benefit from larger-scale validation. Future work will expand evaluation to diverse datasets covering extreme weather and safety-critical scenarios, explore adaptive temporal sampling, and investigate domain-specific fine-tuning.

\bibliographystyle{IEEEtran}
\bibliography{references}

\end{document}